\documentclass[conference]{IEEEtran}
\IEEEoverridecommandlockouts                                
\usepackage{cite}
\usepackage{amsmath,amssymb,amsfonts}
\usepackage{algorithmic}
\usepackage{graphicx}
\usepackage{textcomp}
\usepackage{color}
\usepackage{bm}
\usepackage[table]{xcolor}
\usepackage{colortbl}
\usepackage{multirow}

\usepackage[font=small]{caption}
\usepackage[colorlinks,linkcolor=blue,anchorcolor=blue,citecolor=blue]{hyperref}
\def\BibTeX{{\rm B\kern-.05em{\sc i\kern-.025em b}\kern-.08em
    T\kern-.1667em\lower.7ex\hbox{E}\kern-.125emX}}
\begin{document}
% \title{ Dual-Domain Hybrid Encoder and Enhanced Query Selection for Small Object Detection with Detection Transformer
% }
\title{SO-DETR: Leveraging Dual-Domain Features and Knowledge Distillation for Small Object Detection}

\author{\IEEEauthorblockN{Huaxiang Zhang, Hao Zhang, Aoran Mei, Zhongxue Gan* ,Guo-Niu Zhu*}
\IEEEauthorblockA{\textit{Academy for Engineering and Technology, Fudan University, Shanghai 200433, China}\\
\textit{*Corresponding author: ganzhongxue@fudan.edu.cn, guoniu\_zhu@fudan.edu.cn}}}

\maketitle

\begin{abstract}
Detection Transformer-based methods have achieved significant advancements in general object detection. However, challenges remain in effectively detecting small objects. One key difficulty is that existing encoders struggle to efficiently fuse low-level features. Additionally, the query selection strategies are not effectively tailored for small objects. To address these challenges, this paper proposes an efficient model, Small Object Detection Transformer (SO-DETR). The model comprises three key components: a dual-domain hybrid encoder, an enhanced query selection mechanism, and a knowledge distillation strategy. The dual-domain hybrid encoder integrates spatial and frequency domains to fuse multi-scale features effectively. This approach enhances the representation of high-resolution features while maintaining relatively low computational overhead. The enhanced query selection mechanism optimizes query initialization by dynamically selecting high-scoring anchor boxes using expanded IoU, thereby improving the allocation of query resources. Furthermore, by incorporating a lightweight backbone network and implementing a knowledge distillation strategy, we develop an efficient detector for small objects. Experimental results on the VisDrone-2019-DET and UAVVaste datasets demonstrate that SO-DETR outperforms existing methods with similar computational demands. The project page is available at \url{https://github.com/ValiantDiligent/SO_DETR}.
\end{abstract}

%%%%%%%%%%%%%%%%%%%%%%%%%%%%%%%%%%%%%%%%%%%%%%%%%%%%%%%%%%%%%%%%%%%%%%%%%%%%%%%%
\section{Introduction}
Object detection is a core task in computer vision that focuses on identifying and localizing objects in images\cite{w1}. Despite significant progress in this field, detecting small objects remains a challenging problem. Small object detection is difficult due to tiny sizes, complex backgrounds, and the need to balance performance with low computational costs \cite{MITTAL2020104046}. To address these challenges, research and development have increasingly focused on deep learning-based object detection methods. These methods are broadly categorized into Convolutional Neural Network (CNN)-based approaches and Transformer-based approaches. Among CNN-based methods, the YOLO detector is the most renowned due to its reasonable trade-off between speed and accuracy\cite{w1}. However, these methods often rely on manually designed components, such as anchor boxes generated based on human expertise and non-maximum suppression (NMS) as a post-processing step \cite{TONG2020103910}.
\begin{figure}[ht]
\centering
\includegraphics[width=8.5cm]{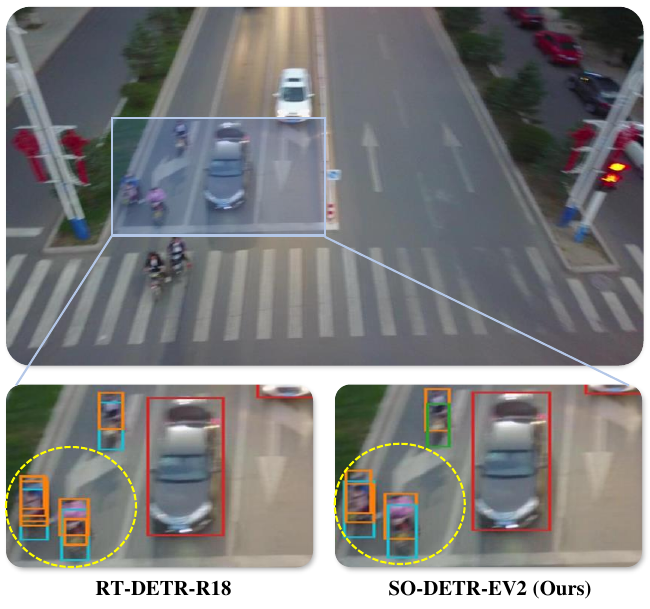}
\caption{Comparison of detection results between RT-DETR-R18 and our proposed SO-DETR-EV2. 
Within the yellow circle, our model outputs fewer overlapping bounding boxes.}
\label{figure_1}
\end{figure}

Detection Transformer (DETR) offers a new object detection method that does not require traditional steps such as anchor boxes and NMS \cite{detr}. Nevertheless, DETR is not efficient, and its performance in small object detection tasks remains suboptimal. Deformable-DETR \cite{zhudeformable} incorporates deformable attention mechanisms. This enhancement improves the detection capability for small-sized objects by utilizing multi-scale feature maps. DINO \cite{zhangdino} further improves the model's detection performance. However, these methods suffer from large computational demands and poor real-time performance. RT-DETR \cite{Zhao_2024_CVPR} achieves a balance between speed and accuracy by decoupling encoder. 
As illustrated in Fig.~\ref{figure_1}, the model is primarily designed for natural images, which presents challenges when applied to detecting small objects.

To balance speed and accuracy in small object detection, we enhance our model by integrating low-level features from dual domains. Additionally, we optimize query selection specifically for small objects. Furthermore, we develop a lightweight and efficient small object detector using knowledge distillation methods.
\begin{figure*}[ht]
\centering
\includegraphics[width=17.2cm]{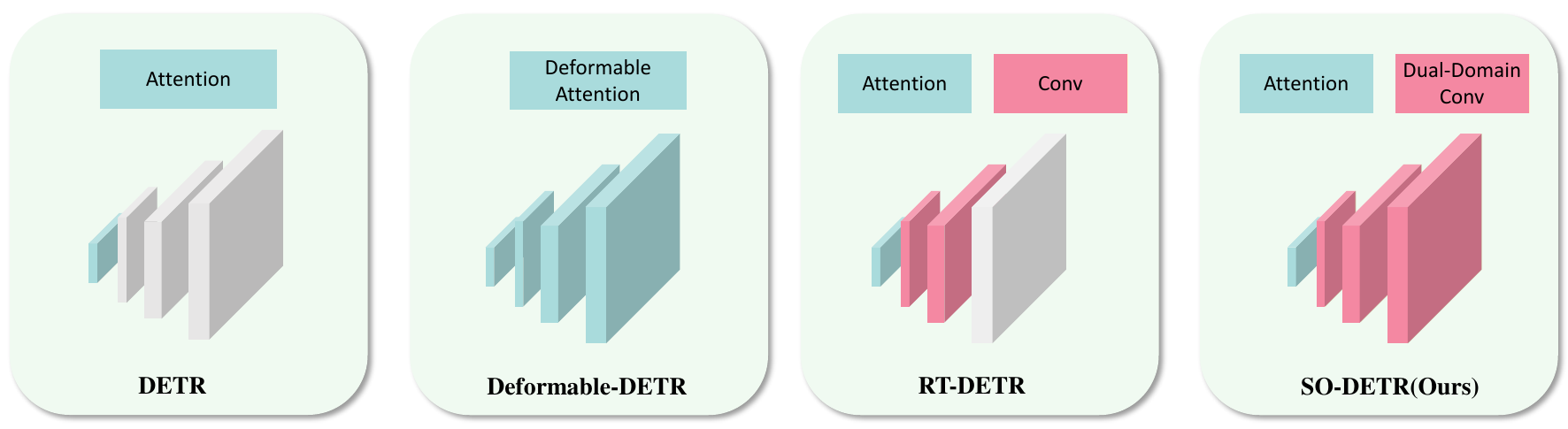}
\caption{Evolution of encoder architectures in DETR-based models and their encoder mechanisms. All models utilize feature layers extracted by a backbone network as input.  } 
\label{RelatedWork}
\end{figure*}

Our main contributions are summarized as follows.
\begin{enumerate}
\item We introduce a dual-domain hybrid encoder that integrates spatial and frequency features. This encoder enhances high-resolution feature representation while maintaining low computational cost.
\item We propose an improved query selection method that dynamically identifies high-scoring anchor boxes as queries using expanded IoU. 
\item We design a knowledge distillation strategy tailored for DETR-like models and integrate it with a lightweight backbone network. This combination results in an efficient and lightweight detector specifically optimized for small object detection. 
\end{enumerate}

\section{Related Work}
\subsection{Detection Transformer}
DETR\cite{detr} uses the transformer architecture to create an end-to-end object detector. Unlike traditional methods, DETR treats object detection as a set prediction problem with bipartite matching. As shown in Fig. \ref{RelatedWork}, DETR feeds the highest-level backbone features into the transformer encoder for self-attention computation. However, DETR performs poorly in detecting small objects due to limited access to lower-level information. Deformable-DETR\cite{zhudeformable} improves small object detection by using deformable attention, which is computationally efficient and applies self-attention to four feature maps from backbone. For better real-time performance, RT-DETR \cite{Zhao_2024_CVPR} only performs self-attention on the highest-level features and uses convolutional structures for cross-scale fusion. 

Many studies have also explored the query selection. Conditional DETR \cite{meng2021-CondDETR} introduces 256-dimensional learnable vectors to define $(x, y)$ reference points and incorporates positional information into the transformer decoder. DAB-DETR \cite{liu2022dabdetr} introduces 256-dimensional vectors to represent $(x, y, w, h)$ reference points. DN-DETR \cite{li2022dn} addresses long training times caused by ambiguity in Hungarian matching by introducing query denoising. DINO \cite{zhangdino} selects top-$k$ features based on classification scores from the encoder. RT-DETR enhances query selection by considering IoU.

Our proposed Small Object Detection Transformer (SO-DETR) enhances the encoder by adding a dual-domain fusion module. This module efficiently fuses feature maps with four-times downsampling, which is crucial for small object detection while keeping computational costs low. Additionally, we incorporate an IoU-aware query selection mechanism using Expanded-IoU, which dynamically selects queries that are more favorable for small object detection.
\begin{figure*}[ht]
\centering
\includegraphics[width=17.5cm]{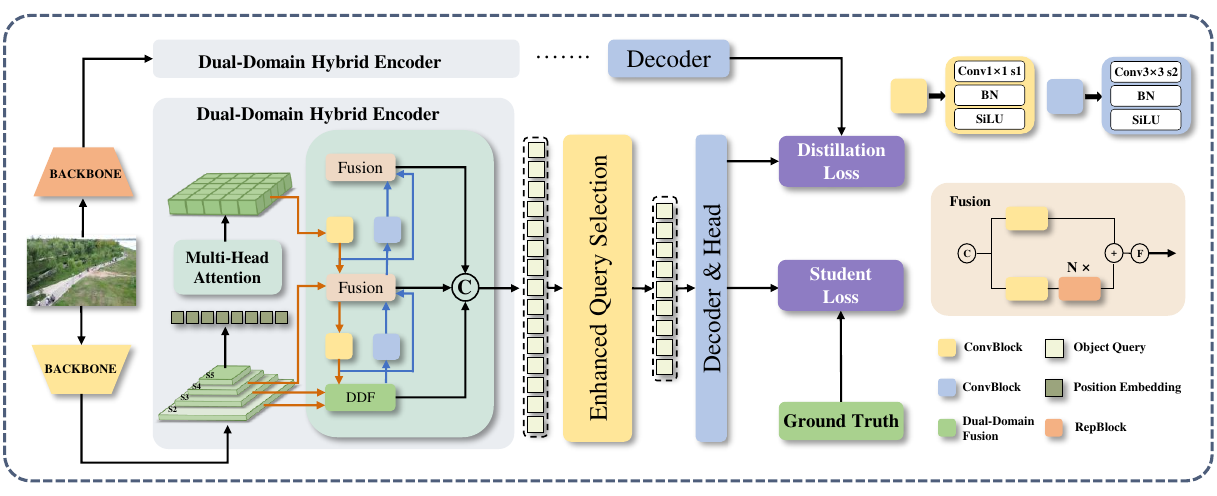}
\caption{Overview of the SO-DETR architecture with knowledge distillation. Multi-scale features from the backbone’s four stages are fed into the dual-domain hybrid encoder, which transforms them into a sequence of image features. The enhanced query selection module then selects a fixed number of these features as initial object queries for the decoder. The decoder, with auxiliary prediction heads, iteratively refines the queries to predict object categories and bounding boxes. During knowledge distillation, the output of the teacher model’s decoder is used to compute the distillation loss with respect to the student model’s predictions.} 
\label{figure_overview}
\end{figure*}

\begin{figure}[ht]
\centering
\includegraphics [width=8.6cm]{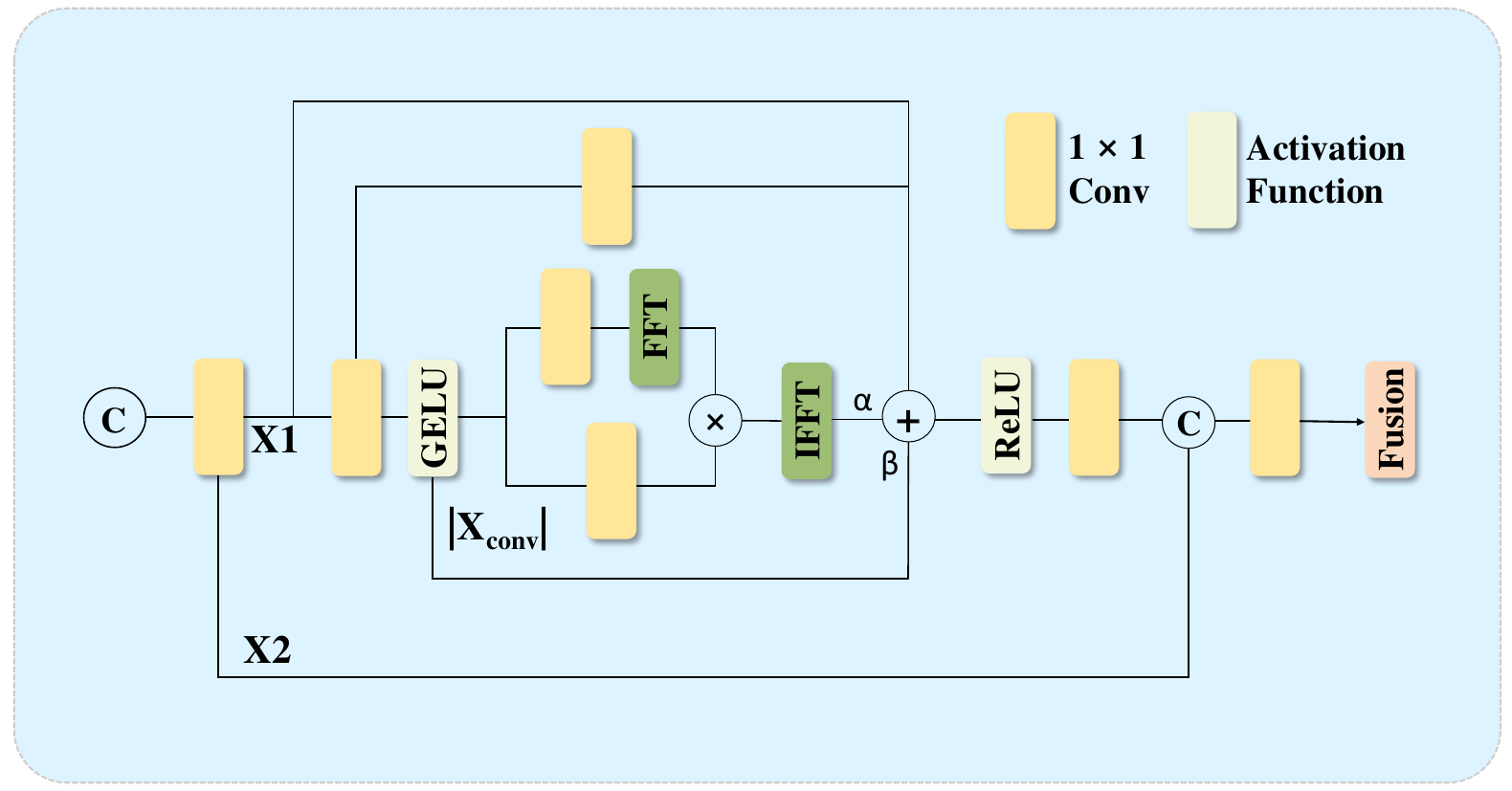}
\caption{The Dual-Domain Fusion block in encoder. FFT and IFFT denote the Fast Fourier Transform and Inverse Fast Fourier Transform, respectively.}
\label{figure_msff}
\end{figure}

\subsection{Knowledge Distillation }
Knowledge Distillation (KD) is a technique for transferring knowledge from a large teacher model to a smaller student model \cite{hinton2015distilling}. In object detection, KD methods typically involve transferring knowledge from the backbone, neck, and detection head of the teacher model to the student model \cite{chen2017learning}. However, most existing KD approaches are designed for CNN-based architectures and are not directly applicable to Transformer-based models like DETR due to architectural differences.

Recent works \cite{chang2023detrdistill, wang2024kd} have addressed this limitation by introducing response-based and feature-based distillation specifically tailored for DETR, as well as specialized object queries to decouple detection from distillation tasks. These adaptations enable effective knowledge transfer in Transformer-based object detectors.

Compared to existing methods, our method focuses on enhancing small object detection within DETR-like models. We concentrate on the outputs of the last decoder layer to compute classification and localization distillation losses.
\section{Methodology}
As shown in Fig. \ref{figure_overview}, this study proposes the SO-DETR model, which is built upon the architecture of RT-DETR \cite{Zhao_2024_CVPR}. Our approach comprises three key components: a dual-domain hybrid encoder, an enhanced query selection mechanism, and a knowledge distillation strategy.

\subsection{Dual-Domain Hybrid Encoder}
High-level features effectively capture the relationships between different conceptual entities due to their richer semantic information. Lower-level features contribute less effectively to attention operations. Therefore, our dual-domain hybrid encoder applies self-attention operations exclusively to the \(\bm{S_5}\) layer. When performing cross-scale feature fusion, we combine the \(\bm{S_2}\) and \(\bm{S_3}\) layers using Dual-Domain Fusion module (DDF).

Fig.~\ref{figure_msff} illustrates the architectural of DDF. Given an input feature map \(\mathbf{X} \in \mathbb{R}^{C \times H \times W}\), the module begins with a convolutional operation followed by channel-wise splitting:
\begin{equation}
[\mathbf{X}_1, \mathbf{X}_2] = \text{Split}(\text{Conv}(\mathbf{X}), [C_1, C_2], \text{dim}=0)
\label{eq:split}
\end{equation}

Here, \(\text{Conv}(\cdot)\) denotes a convolutional layer that maintains the spatial dimensions while adjusting the channel dimensions. The split divides the convolved features into two distinct branches: \(\mathbf{X}_1\) with \(C_1\) channels and \(\mathbf{X}_2\) with \(C_2\) channels, where \(C_1 + C_2 = C\). This bifurcation allows for specialized transformation in subsequent stages. Specifically, \(C_1\) is set to be one-quarter of the total number of channels.

The first branch, \(\mathbf{X}_1\), undergoes further convolution followed by a Gaussian Error Linear Unit (GELU) activation to introduce non-linearity:
\begin{equation}
\mathbf{X}_{\text{conv}} = \text{GELU}(\text{Conv}(\mathbf{X}_1))
\label{eq:conv_gelu}
\end{equation}

The GELU activation is chosen for its smooth non-linear properties. To capture intricate feature dependencies, the convolved features \(\mathbf{X}_{\text{conv}}\) are processed in the frequency domain as follows:
\begin{equation}
\begin{aligned}
\mathbf{X}_{\text{out}} &= \alpha_1 \cdot \left|\text{IFFT}\left(\text{FFT}\left(\text{Conv}\left(|\mathbf{X}_{\text{conv}}|\right)\right)\right) \cdot \text{Conv}\left(|\mathbf{X}_{\text{conv}}|\right)\right| \\
&+\text{Conv}\left(\text{ReLU}\left(\mathbf{X}_1 + \text{Conv}\left(\mathbf{X}_{\text{conv}}\right) +\beta_1 \cdot |\mathbf{X}_{\text{conv}}|\right)\right)
\end{aligned}
\label{eq:xout}
\end{equation}

The coefficients \(\alpha_1\) and \(\beta_1\) are learnable parameters that balance the contributions from the frequency-enhanced features and the spatial residual connections, respectively. The outputs from both branches are concatenated and passed through a final convolutional layer to integrate the refined features.
\begin{equation}
\mathbf{X}_{\text{final}} = \text{Conv}([\mathbf{X}_{\text{out}}, \mathbf{X}_2])
\label{eq:final}
\end{equation}

This concatenation ensures that both frequency-enhanced and original features contribute to the final representation. The resulting vectors are subsequently fed into a Fusion block\cite{Zhao_2024_CVPR}. 

\subsection{Enhanced Query Selection}
In SO-DETR, anchor boxes are generated on multi-scale feature maps using a fixed grid. They are subsequently refined through position transformations and selection in logit space. During training, the regression head dynamically adjusts the anchor boxes to match the target's position and scale. To improve small object detection, a classification head incorporates the expanded IoU score to dynamically select the top-$k$ high-scoring anchor boxes as queries.

The predicted bounding box \( B_p \) and the ground truth bounding box \( B_{gt} \) represent the model's prediction and the actual target, respectively. Expanded-IoU is defined as the Intersection over Union between the scaled predicted bounding box \( B_p' \) and the scaled ground truth bounding box \( B_{gt}' \). Both boxes are proportionally scaled by a factor \( (\alpha_2 > 1) \) while maintaining their centers fixed. 
This scaling operation increases both the width and height of the bounding boxes.
The scaled bounding boxes are defined as \( B_p' = (x_p, y_p, w_p \times \alpha_2, h_p \times \alpha_2) \) and \( B_{gt}' = (x_{gt}, y_{gt}, w_{gt} \times \alpha_2, h_{gt} \times \alpha_2) \), where \( (x_p, y_p) \) and \( (x_{gt}, y_{gt}) \) are the centers, and \( w_p, h_p, w_{gt}, h_{gt} \) represent the widths and heights of the predicted and ground truth boxes, respectively. The Expanded-IoU is defined as:
\begin{equation} 
\text{Expanded-IoU}(B_p, B_{gt}) = \left|B_p' \cap B_{gt}'\right| / \left|B_p' \cup B_{gt}'\right|.
\end{equation}

In our framework, we utilize Expanded-IoU to compute both the classification loss and the IoU loss, while the SIoU\cite{gevorgyan2022siou} is used specifically for calculating the IoU loss. The Expanded-SIoU is then defined as:
\begin{equation}
\text{Expanded-SIoU} = \text{SIoU} - \text{IoU} + \text{Expanded-IoU}.
\end{equation}

\subsection{Knowledge Distillation for Small Object Detection}
We develop a teacher model based on RT-DETR-R50 and a lightweight student model derived from RT-DETR-R18 by replacing the traditional ResNet-18 backbone with EfficientFormerV2\cite{li2022rethinking}. Both models incorporate our proposed enhanced query selection and dual-domain hybrid encoder methods to improve small object detection capabilities. To further enhance the student model's performance, we employ a knowledge distillation strategy where the student learns from the outputs of the teacher model’s final decoder layer. The design of the loss functions used in the distillation process is described below.  

We employ Binary Cross-Entropy (BCE) to measure the discrepancy between the classification scores of the student model and the teacher model. This loss term quantifies the deviation between the student’s classification predictions and those of the teacher. It encourages the student model to achieve more accurate classification.
\begin{equation} 
\mathcal{L}_c = \text{BCE}(s_c, t_c)
\end{equation}
where \( s_c \) and \( t_c \) represent the classification score outputs from the final decoder layer of the student model and the teacher model, respectively.

We utilize the L1 loss to compute the differences in bounding box coordinates, weighted by the object confidence scores from the teacher model. This loss ensures the student model approximates the teacher’s bounding box predictions, especially for high-confidence objects.
\begin{equation} 
\mathcal{L}_{L1} = \text{L1}(s_b, t_b) \times t_o
\end{equation}
where \( s_b \) and \( t_b \) denote the bounding box coordinates predicted by the student model and the teacher model, respectively, and \( t_o \) represents the object confidence score from the teacher model. To enhance small object detection, we apply Expanded-SIoU. By introducing Expanded- SIoU, IoU loss  helps the student model align the bounding boxes of small objects.
\begin{equation} 
\mathcal{L}_{IoU} = \left(1 - \text{Expanded-SIoU}(s_b, t_b)\right)
\end{equation}

The total distillation loss is formulated as follows:
\begin{equation} 
\mathcal{L}_{total} = \alpha \mathcal{L}_c + \beta \mathcal{L}_{L1} + \gamma \mathcal{L}_{IoU}
\end{equation}
where \( \alpha \), \( \beta \), and \( \gamma \) are constant coefficients that balance the contributions of each loss component.

\begin{table*}[htbp]
  \caption{Experimental Results on the VisDrone-2019-DET Dataset}
  \begin{center}
  \footnotesize
  \begin{tabular}{l l c c c c c c} % Columns: Model, Backbone, Publication, InputSize, Params(M), GFLOPs, AP, AP$_{50}$
  \hline
    \textbf{Model} & \textbf{Backbone} & \textbf{Publication} & \textbf{Input Size} & \textbf{Params (M)} & \textbf{GFLOPs} & \textbf{AP} & \textbf{AP$_{50}$} \\
    \hline
    \multicolumn{8}{l}{\cellcolor{gray!20}\textit{Low Computation }} \\
    YOLOv11-S\cite{khanam2024yolov11} & CSPDarknet (C3K2, C2PSA) & arXiv2024 & 640$\times$640 & 9.4 & 21.3 & 23.0 & 38.7 \\
    YOLOv10-S\cite{wang2024yolov10} & CSPDarknet (SCDown) & arXiv2024 & 640$\times$640 & 8.0 & 24.5 & 22.2 & 37.4 \\
    YOLOv9-S\cite{wang2024yolov9} & CSPDarknet (GELAN, PGI) & ECCV2025 & 640$\times$640 & 7.2 & 26.7 & 22.7 & 38.3 \\
    YOLOv8-S\cite{yolov8_ultralytics} & CSPDarknet (C2F) & - & 640$\times$640 & 11.1 & 28.5 & 22.3 & 37.6 \\
    HIC-YOLOv5\cite{hicyolo} & CSPDarknet & ICRA2024 & 640$\times$640 & 9.4 & 31.2 & 20.8 & 36.1 \\
    SO-DETR (Ours) & EfficientFormerV2 & - & 640$\times$640 & 12.1 & 33.3 & 28.2 & 46.7 \\
    SO-DETR (Distilled) & EfficientFormerV2 & - & 640$\times$640 & 12.1 & 33.3 & \textbf{28.8} & \textbf{47.5} \\
    \hline
    \multicolumn{8}{l}{\cellcolor{gray!20}\textit{Medium Computation }} \\
    YOLOv10-M\cite{wang2024yolov10} & CSPDarknet (SCDown) & arXiv2024 & 640$\times$640 & 15.4 & 59.1 & 24.5 & 40.5 \\
    
    YOLOv11-M\cite{khanam2024yolov11} & CSPDarknet (C3K2, C2PSA) & arXiv2024 & 640$\times$640 & 20.0 & 67.7 & 25.9 & 43.1 \\
    YOLOv9-M\cite{wang2024yolov9} & CSPDarknet (GELAN, PGI) & ECCV2025 & 640$\times$640 & 20.1 & 76.8 & 25.2 & 42.0 \\
    
    YOLOv8-M\cite{yolov8_ultralytics} & CSPDarknet (C2F) & - & 640$\times$640 & 25.9 & 78.9 & 24.6 & 40.7 \\
    UAV-DETR-R18\cite{zhang2025uavdetrefficientendtoendobject} & ResNet18 & arXiv2025 & 640$\times$640 & 20.0 & 77.0 & 29.8 & 48.8 \\
    RT-DETR-R18\cite{Zhao_2024_CVPR} & ResNet18 & CVPR2024 & 640$\times$640 & 20.0 & 57.3 & 26.7 & 44.6 \\
    SO-DETR (Ours) & ResNet18 & - & 640$\times$640 & 20.5 & 64.3 & \textbf{29.9} & \textbf{49.0} \\
    \hline
    \multicolumn{8}{l}{\cellcolor{gray!20}\textit{High Computation}} \\
    YOLOv10-L\cite{wang2024yolov10} & CSPDarknet (SCDown) & arXiv2024 & 640$\times$640 & 24.4 & 120.3 & 26.3 & 43.1 \\
    PP-YOLOE-P2-Alpha-l\cite{ppdet2019} & CSPRepResNet & - & 640$\times$640 & 54.1 & 111.4 & 30.1 & 48.9 \\
    YOLOv8-L\cite{yolov8_ultralytics} & CSPDarknet (C2F) & - & 640$\times$640 & 43.7 & 165.2 & 26.1 & 42.7 \\
    DetectoRS w/ RFLA\cite{xu2022rfla} & ResNet50 & ECCV2022 & 800$\times$800 & 123.2 & 160.0 & 27.4 & 45.3 \\
    DCFL\cite{Xu_2023_CVPR} & ResNet50 & CVPR2023 & 1024$\times$1024 & 36.1 & 157.8 & - & 32.1 \\
    Deformable DETR\cite{zhudeformable} & ResNet50 & ICLR2020 & 1333$\times$800 & 40.0 & 173.0 & 27.1 & 42.2 \\
    DETR\cite{detr} & ResNet50 & ECCV2020 & 1333$\times$750 & 60.0 & 187.0 & 24.1 & 40.1 \\
    YOLOv11-X\cite{khanam2024yolov11} & CSPDarknet (C3K2, C2PSA) & arXiv2024 & 640$\times$640 & 56.8 & 194.5 & 28.1 & 45.6 \\
    QueryDet\cite{Yang_2022_CVPR} & ResNet50 & CVPR2022 & 2400$\times$2400 & 33.9 & 212.0 & 28.3 & 48.1 \\
    ClusDet\cite{yang2019clustered} & ResNet50 & ICCV2019 & 1000$\times$600 & 30.2 & 207.0 & 26.7 & 50.6 \\
    UAV-DETR-R50\cite{zhang2025uavdetrefficientendtoendobject} & ResNet50 & arXiv2025 & 640$\times$640 & 42.0 & 170.0 & \textbf{31.5} & 51.1 \\
    RT-DETR-R50\cite{Zhao_2024_CVPR} & ResNet50 & CVPR2024 & 640$\times$640 & 42.0 & 129.9 & 28.4 & 47.0 \\
    SO-DETR (Ours) & ResNet50 & - & 640$\times$640 & 44.4 & 161.4 & \textbf{31.5} & \textbf{51.5} \\
    \hline
  \end{tabular}%
  \end{center}
  \label{table:detectors_reformatted}
\end{table*}
\begin{table}[htbp]
\caption{Experimental Results on UAVVaste Dataset}
\centering
\footnotesize
\begin{tabular}{l c c c c}
\hline
\textbf{Model} & \textbf{Params (M)} & \textbf{GFLOPs} & \textbf{AP} & \textbf{AP$_{50}$} \\
\hline
YOLOv11-S\cite{khanam2024yolov11} & 9.4 & 21.3 & 27.8 & 63.0\\
HIC-YOLOv5\cite{hicyolo} & 9.4 & 31.2 & 30.5 & 65.1\\
UAV-DETR-EV2\cite{zhang2025uavdetrefficientendtoendobject}& 13 & 43 & 37.1 & 70.6\\
RT-DETR-R18\cite{Zhao_2024_CVPR} & 20.0 & 57.3 & 36.3 & 72.6\\
RT-DETR-R50\cite{Zhao_2024_CVPR} & 42.0 & 129.9 & 37.4 & 73.5\\
\hline
SO-DETR-R50 (Ours)& 44.4 & 161.4 & 37.5& 76.4\\
SO-DETR-R18 (Ours)& 20.5 & 64.3 & 35.1& 72.1\\
SO-DETR-EV2 (Ours)& 12.1 & 33.3 & 33.7& 70.6\\
SO-DETR-EV2 (Distilled) & 12.1 & 33.3 & 36.9 & 73.6 \\
\hline
\end{tabular}
\label{table_vaste}
\end{table}

Finally, we implement a linear decay schedule for the distillation loss. This scheme assigns higher weights to the distillation loss early in training. As training progresses, the weight of the distillation loss is gradually reduced.

\section{Experiments}
\subsection{Experimental Setup}
\textbf{Datasets:}
To emphasize the model's ability to detect small targets, we use datasets with a high prevalence of such objects. Comprehensive quantitative evaluations are conducted on two object detection benchmarks: VisDrone \cite{zhu2021detection} and UAVVaste \cite{rs13050965}.

The VisDrone-2019-DET dataset contains 6471 training images, 548 validation images, and 3190 test images. These images were captured by UAVs operating at different altitudes and across various locations. Each image was annotated with bounding boxes for ten object categories: pedestrian, person, car, van, bus, truck, motorbike, bicycle, awning-tricycle, and tricycle. Since the test set is not publicly available, we followed the evaluation setup from previous studies\cite{Yang_2022_CVPR,Xu_2023_CVPR}. For our experiments, we used the training set for model training and the validation set for performance evaluation.

Furthermore, we train the SO-DETR network using the UAVVaste dataset to assess the model's generalization capabilities across different datasets. UAVVaste is specifically designed for aerial rubbish detection and comprises 772 images with 3716 manually labeled waste annotations in both urban and natural environments. The dataset was created to address the lack of domain-specific data for waste detection using drones or UAVs. We employ the training subset for training and the test subset for evaluating detection performance.

\textbf{Implementation Details:}
All models are trained on NVIDIA GeForce RTX 3090. The experimental configuration adheres to the default parameters from the official implementations, with training conducted until convergence. Our approach is implemented and trained for 350 epochs using a batch size of 4, incorporating an early stopping mechanism with a patience value of 40. For the distillation experiments, training is conducted for 600 epochs. We use the same data augmentation strategy as \cite{zhang2025uavdetrefficientendtoendobject}.

We report the standard COCO metrics, including AP (averaged over uniformly sampled IoU thresholds ranging from 0.50 to 0.95 with a step size of 0.05), and AP$_{50}$ (AP at an IoU threshold of 0.50). We also report the performance of the model across different object sizes using AP$_{\text{small}}$, AP$_{\text{medium}}$, and AP$_{\text{large}}$. These metrics specifically focus on the detection performance of small, medium, and large objects, respectively. Objects are classified as small, medium, or large based on their area in the image. Small objects have an area of less than \( 32^2 \) pixels, medium objects range from \( 32^2 \) to \( 96^2 \) pixels, and large objects have an area greater than \( 96^2 \) pixels in area.

\subsection{Comparative Experiments} 
As shown in Table \ref{table:detectors_reformatted}, we evaluate the performance of our proposed SO-DETR models against existing object detectors and small object detectors using the VisDrone-2019-DET dataset. 
The models are categorized into three groups based on their computational complexity: Low Computation for models with less than 50 GFLOPs, Medium Computation for models with 50 to 100 GFLOPs, and High Computation for models with more than 100 GFLOPs.

\textbf{Low Computation:}  
In the low computation category, SO-DETR-EV2 achieves an AP of 28.2\% and an AP$_{50}$ of 46.7\%, outperforming other real-time detectors such as YOLOv11-S \cite{khanam2024yolov11} (23.0\% AP, 38.7\% AP$_{50}$), YOLOv10-S\cite{wang2024yolov10} (22.2\% AP, 37.4\% AP$_{50}$), and HIC-YOLOv5 \cite{hicyolo} (20.8\% AP, 36.1\% AP$_{50}$). The distilled SO-DETR-EV2 variant further enhances performance with an AP of 28.8\% and an AP$_{50}$ of 47.5\%.

\textbf{Medium Computation :}  
Within the medium computation range, SO-DETR-R18 demonstrates superior accuracy with an AP of 29.9\% and an AP$_{50}$ of 49.0\%, surpassing YOLOv10-M (24.5\% AP, 40.5\% AP$_{50}$), YOLOv11-M (25.9\% AP, 43.1\% AP$_{50}$), RT-DETR-R18 \cite{Zhao_2024_CVPR} (26.7\% AP, 44.6\% AP$_{50}$), and UAV-DETR-R18\cite{zhang2025uavdetrefficientendtoendobject} (29.8\% AP, 48.8\% AP$_{50}$).
\begin{table*}[htbp]
\caption{Performance comparison with different configurations.}
\centering
\footnotesize
\begin{tabular}{cccccccccc} % 定义10列布局
\hline
\textbf{Backbone} & \textbf{Query Select} & \textbf{Encoder} & \textbf{Params(M)} & \textbf{GFLOPs} & \textbf{AP} & \textbf{AP$_{50}$} & \textbf{AP$_{Small}$} & \textbf{AP$_{Medium}$} & \textbf{AP$_{Large}$} \\
\hline
\multirow{4}{*}{R50} &      &      & 42.0 & 129.9 & 30.6 & 49.8 & 21.7 & 42.1 & \textbf{53.2} \\
                     & \checkmark &      & 42.0 & 129.9 & 31.0 & 50.4 & 21.9 & 42.5 & 49.9 \\
                     &      & \checkmark & 44.4 & 161.4 & 30.9 & 50.9 & 22.3 & 42.3 & 49.7 \\
                     & \checkmark & \checkmark & 44.4 & 161.4 & \textbf{31.5} & \textbf{51.5} & \textbf{22.4} & \textbf{43.5} & 46.7 \\
\hline
\multirow{4}{*}{R18} &      &      & 20.0 & 57.3 & 27.6 & 45.8 & 19.3 & 37.9 & \textbf{43.4} \\
                     & \checkmark &      & 20.0 & 57.3 & 27.7 & 46.3 & 19.4 & 38.1 & 40.9 \\
                     &      & \checkmark & 20.5 & 64.3 & 28.9 & 47.4 & 20.6 & 39.7 & 45.4 \\

                     & \checkmark & \checkmark & 20.5 & 64.3 & \textbf{29.9} & \textbf{49.0} & \textbf{21.1} & \textbf{41.2} & 42.9 \\
\hline
\multirow{4}{*}{EV2} &      &      & 11.9 & 29.8 & 25.5 & 43.2 & 17.1 & 35.7 & 42.5 \\
                     & \checkmark &      & 11.9 & 29.8 & 26.7 & 45.2 & 18.6 & 36.6 & \textbf{44.3} \\
                     &      & \checkmark & 12.1 & 33.3 & 26.4 & 44.5 & 18.5 & 36.3 & 41.9 \\
                     & \checkmark & \checkmark & 12.1 & 33.3 & \textbf{28.2} & \textbf{46.7} & \textbf{19.7} & \textbf{38.2} & 43.1 \\
\hline
\end{tabular}
\label{tab:performance_comparison}
\end{table*}
\begin{table*}[htbp]
\caption{Effect of Combining Different Distillation Strategies}
\centering
\footnotesize
% \resizebox{\textwidth}{!}{ % 让表格自适应页面宽度
\begin{tabular}{l c c c c c c c} % 定义列布局
\hline
\textbf{Model} & \textbf{Params(M)} & \textbf{GFLOPs} & \textbf{AP} & \textbf{AP$_{50}$} & \textbf{AP$_{Small}$} & \textbf{AP$_{Medium}$} & \textbf{AP$_{Large}$} \\
\hline
SO-DETR-R50(Teacher) & 42.0 & 161.4 & 31.5 & 51.5 & 22.4 & 43.5 & 46.7 \\ %ok
SO-DETR-EV2(Student) & 12.1 & 33.3 & 28.2 & 46.7 & 19.7 & 38.2 & 43.1 \\  %ok
+Constant + GIoU & 12.1 & 33.3 & 28.3 & 46.7 & 19.8 & 38.7 & 43.0 \\
+Constant + Expanded-SIoU & 12.1 & 33.3& 27.4 & 45.7 & 18.9 & 38.3 & 41.9 \\
+Cosine + GIoU & 12.1 & 33.3 & 28.1 & 46.7 & 19.5 & 38.8 & 40.3 \\
+Cosine + Expanded-SIoU & 12.1 & 33.3 & 28.0 & 46.2 & 19.8 & 38.4 & 41.5 \\
+Linear + GIoU & 12.1 & 33.3 & \textbf{28.8} & 47.2 & 20.0 & \textbf{39.8} & \textbf{43.2} \\
+Linear + Expanded-SIoU & 12 & 33 & \textbf{28.8} & \textbf{47.5} & \textbf{20.5} & 39.1 & \textbf{43.2} \\
\hline
\end{tabular}
% }
\label{table:KDdetec}
\end{table*} 

\textbf{High Computation :}  
SO-DETR-R50 achieves an AP of 31.5\% and an AP$_{50}$ of 51.5\%, outperforming UAV-DETR-R50\cite{zhang2025uavdetrefficientendtoendobject} (31.5\% AP, 51.1\% AP$_{50}$), YOLOv10-L (26.3\% AP, 43.1\% AP$_{50}$), PP-YOLOE-P2-Alpha-l \cite{ppdet2019} (30.1\% AP, 48.9\% AP$_{50}$), YOLOv11-X (28.1\% AP, 45.6\% AP$_{50}$), QueryDet \cite{Yang_2022_CVPR} (28.3\% AP, 48.1\% AP$_{50}$), ClusDet \cite{yang2019clustered} (26.7\% AP, 50.6\% AP$_{50}$), and RT-DETR-R50 (28.4\% AP, 47.0\% AP$_{50}$). 

As shown in Table \ref{table:detectors_reformatted}, our SO-DETR models consistently achieve the highest AP and AP\(_{50}\) scores across low, medium, and high computational complexity categories on the VisDrone-2019-DET dataset. 

To further validate our method, we conduct experiments on the UAVVaste dataset. As shown in Table \ref{table_vaste}, SO-DETR-R50 achieves an AP of 37.5\% and an AP$_{50}$ of 76.4\%, compared to RT-DETR-R50's AP of 37.4\% and AP$_{50}$ of 73.5\%. The distilled SO-DETR-EV2 variant obtains an AP of 36.9\% and an AP$_{50}$ of 73.6\%, compared to the non-distilled SO-DETR-EV2's AP of 33.7\% and AP$_{50}$ of 70.6\%.

These results demonstrate that SO-DETR effectively enhances small object detection while maintaining competitive computational efficiency. Additionally, our SO-DETR models exhibit superior performance and robustness across various datasets.

\subsection{Ablation Studies}
To further validate the effectiveness of our method, we conduct ablation experiments on the VisDrone dataset, as presented in Table \ref{tab:performance_comparison}. The experiments are performed using three different backbone networks: ResNet50 (R50), ResNet18 (R18), and EfficientFormerV2 (EV2). For each backbone, we evaluate the impact of integrating the dual-domain hybrid encoder and enhanced query selection modules individually and in combination.

Compared to the R18 backbone, the EV2 backbone significantly reduces computational cost, reaching only 52\% of the original values. This reduction leads to a decrease of 2.1\% in AP and 2.6\% in AP$_{50}$. After introducing the enhanced query selection, AP improves by 1.2\% and AP$_{50}$ increases by 2.0\%. The addition of the dual-domain hybrid encoder method further contributes to AP and AP$_{50}$ gains of 0.9\% and 1.3\%, respectively. When both methods are applied together, the model achieves its best performance. The AP increases by 2.7\% to reach 28.2\%, and the AP$_{50}$ improves by 3.5\%, reaching 46.7\%.

Further ablation studies on the R50 and R18 backbones reveal similar trends. For the R50 model, the enhanced query selection and dual-domain hybrid encoder methods individually increase AP by 0.4\% and 0.3\%, respectively. When both methods are combined, AP improves by 0.9\%, accompanied by gains in AP$_{small}$ of 0.7\% and AP$_{medium}$ of 1.4\%. In the case of the R18 backbone, the enhanced query selection and dual-domain hybrid encoder methods result in AP increases of 0.1\% and 1.3\%, respectively. When both are used together, AP increases by 2.3\%, with AP$_{small}$ and AP$_{medium}$ improving by 1.8\% and 3.3\%, respectively.

We conduct ablation experiments on our knowledge distillation strategy, comparing three different schemes for the decay of the distillation loss weight: a constant distillation loss weight, a non-linear oscillation of the distillation loss weight based on a cosine function, and a gradual linear decay of the distillation loss weight. Additionally, we evaluate the performance of GIoU and Expanded-SIoU for calculating the IoU loss during the distillation process.

As shown in Table \ref{table:KDdetec}, the best distillation performance is achieved when both Expanded-SIoU and the linearly decaying distillation loss weight are used during training. Under this configuration, the model shows improvements across multiple metrics: AP increases by 0.6\%, AP$_{50}$ rises by 0.8\%, AP$_{small}$ improves by 0.8\%, AP$_{medium}$ goes up by 0.9\%, and AP$_{large}$ increases by 0.1\%.
\begin{figure*}[ht]
\centering
\includegraphics[width=17.5cm]{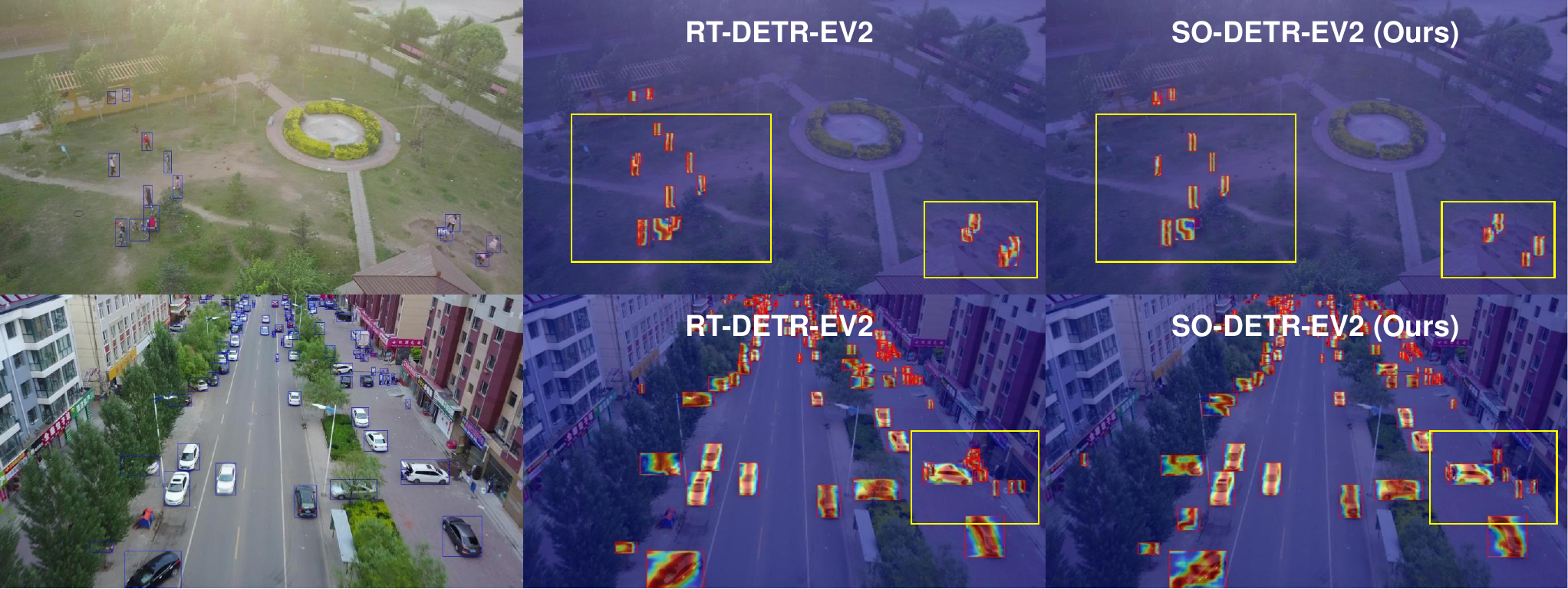}
\caption{Qualitative comparison of detection results and attention heatmaps between RT-DETR-EV2 and our proposed SO-DETR-EV2. Images are from the VisDrone-2019-DET dataset. The yellow boxes highlight areas where our model outperforms RT-DETR-EV2 by generating more precise attention distributions and detecting small and distant objects more effectively. } 
\label{figure_heatmap}
\end{figure*}

\subsection{Visualization}
In Fig. \ref{figure_heatmap}, we present heatmaps overlaid with detection results for small objects from the VisDrone dataset. Compared to the RT-DETR-EV2, the distilled SO-DETR-EV2 model exhibits a significant reduction in false positives, especially in small object detection. Our method demonstrates superior detection accuracy for small objects, highlighting its improved performance in challenging scenarios.

\subsection{Discussion}
From the comparative experimental results, our model achieves the best detection performance under similar parameter counts and computational costs. This superiority holds true when compared to both general object detectors and small object detectors. Specifically, the distilled SO-DETR-EV2 reduces computational load by 41.8\% relative to the default configuration of RT-DETR-R18, while improving AP by 2.1\% and AP$_{50}$ by 2.9\%. To understand the underlying reasons for this performance gain, we analyze the contributions of our proposed components.

A key factor contributing to the enhanced performance is the integration of the dual-domain hybrid encoder. Our dual-domain operations efficiently acquire fine-grained information with low computational overhead. This dual-domain approach enables the model to leverage frequency patterns, which traditional spatial-only encoders might overlook. 

Furthermore, our enhanced query selection mechanism and knowledge distillation strategy improve the model's accuracy without incurring additional computational costs. The enhanced query selection method introduces an enlarged width-height IoU loss function, which significantly accelerates the convergence of small object detection \cite{zhang2023inner}. The enlarged bounding boxes are less sensitive to minor positional shifts, which makes IoU calculations more tolerant. This approach optimizes query resource allocation by prioritizing medium and small objects, enabling the model to detect finer details more effectively. By simultaneously using a linear decay schedule and enlarged bounding boxes during the distillation process, we ensure effective knowledge transfer from the teacher model while allowing the student model to undergo refined optimization as training progresses. This combination of the dual-domain hybrid encoder, enhanced query selection, and knowledge distillation methods enables SO-DETR-EV2 to maintain competitive detection performance even when using a lightweight backbone network.

Moreover, the combined use of the dual-domain hybrid encoder and enhanced query selection methods yields an improvement in experimental performance that surpasses the sum of their individual contributions. In models with the EV2 backbone, the enhanced query selection method alone increases AP by 1.2\%, and the dual-domain hybrid encoder method alone increases AP by 0.9\%. When used in combination, AP increases by 2.7\%. This synergy is consistently demonstrated across all three backbone networks tested. We consider that the dual-domain hybrid encoder provides richer feature representations, while the query selection method ensures effective allocation of query resources. Together, they substantially enhance the model’s performance in detecting medium and small objects.

However, we observe that using R50 and R18 backbones leads to a decrease in AP$_{large}$. This may be because the encoder's focus on high-resolution features reduces the model’s semantic understanding of larger objects, and the query selection method allocates fewer resources to them.

\section{Conclusion}
We present SO-DETR, a Detection Transformer-based model tailored for small object detection. By incorporating a dual-domain hybrid encoder and an enhanced query selection mechanism, SO-DETR effectively combines spatial and frequency features to better capture high-resolution details and optimize query allocation. 
Additionally, our knowledge distillation approach with a lightweight backbone contributes to maintaining computational efficiency. 
Experiments on the VisDrone-2019-DET and UAVVaste datasets demonstrate that SO-DETR achieves competitive accuracy compared to existing methods. Ablation studies confirm the contributions of each component, highlighting the model's capability to address challenges in small object detection within Transformer-based frameworks. Future work will address the decrease in AP$_{large}$ when using our methods by enhancing the balance between high-resolution feature extraction and semantic understanding of larger objects.
\newpage
\bibliographystyle{IEEEtran}
\bibliography{ref}
\end{document}